%% file: paper.tex
\documentclass{article}

\usepackage[preprint,nonatbib]{nips_2018}

\usepackage[square,numbers]{natbib}
\usepackage[utf8]{inputenc} 
\usepackage[T1]{fontenc}    
\usepackage{hyperref}       
\usepackage{url}            
\usepackage{booktabs}       
\usepackage{amsfonts}       
\usepackage{nicefrac}       
\usepackage{microtype}      
\usepackage[utf8]{inputenc}
\usepackage{comment}
\usepackage{color}
\usepackage{array}
\usepackage{multirow}
\usepackage{multicol}
\usepackage{amsmath}
\usepackage{graphicx}
\usepackage{caption}
\usepackage{subfigure} 
\usepackage{float}
\usepackage{wrapfig}

\graphicspath{ {figs/} } 
\usepackage{xargs}                      
\usepackage[colorinlistoftodos,prependcaption,textsize=tiny]{todonotes}

\title{A generic framework for \\privacy preserving deep learning}

\author{
  Théo Ryffel\thanks{Member of the OpenMined community} \\
    Imperial College London \\
    \texttt{tr17@ic.ac.uk} \\
    \And
    Andrew Trask$^{*}$ \\
    DeepMind \\ University of Oxford \\
    \texttt{liamtrask@gmail.com} \\
    \And
    Morten Dahl$^{*}$ \\
    \texttt{mortendahlcs@gmail.com} \\
    \And
    Bobby Wagner$^{*}$ \\
    Case Western Reserve University \\
    \texttt{bobbywagner@case.edu} \\
    \And
    Jason Mancuso$^{*}$ \\
    \texttt{jason@manc.us} \\
    \And
    Daniel Rueckert \\
    Imperial College London \\
    \texttt{dr@ic.ac.uk}
    \And
    Jonathan Passerat-Palmbach \\
    Imperial College London \\
    \texttt{jpassera@ic.ac.uk}
}

\begin{document}

\maketitle

\begin{abstract}
We detail a new framework for privacy preserving deep learning and discuss its assets.
The framework puts a premium on ownership and secure processing of data and introduces a valuable representation based on chains of commands and tensors. This abstraction allows one to implement complex privacy preserving constructs such as Federated Learning, Secure Multiparty Computation, and Differential Privacy while still exposing a familiar deep learning API to the end-user. We report early results on the Boston Housing and Pima Indian Diabetes datasets. While the privacy features apart from Differential Privacy do not impact the prediction accuracy, the current implementation of the framework introduces a significant overhead in performance, which will be addressed at a later stage of the development.
We believe this work is an important milestone introducing the first reliable, general framework for privacy preserving deep learning.
\end{abstract}

\input{sections/introduction}

\input{sections/framework}
\input{sections/mpctensor}
\input{sections/results}
\input{sections/conclusion}
\section{Acknowledgements}
\vspace{-2px}
We would like to address a special thank to the very many contributors to the \href{https://github.com/OpenMined/PySyft/}{OpenMined project PySyft}. In particular, we are very grateful to Parsa Alamzadeh for the development of fixed precision and numerous bug resolutions, Yann Dupis for his integration of PATE and intense cleaning across the repo, Iker Ceballos who worked on federated averaging, Luis Fernando Leal and Kerat for their regular and precious contributions and Abdulrahman Mazhar who introduced gradual typing. Their combined commitment to improving PySyft and to adding new features was decisive. We would all like to thank those who have taken care of keeping the repo clean and documented: Adarsh Kumar, Amit Rastogi, Abhinav Dadhich, Jemery Jordan, Alexandre Granzer-Guay, Hrishikesh Kamath, Sharat Patil and in general \href{https://github.com/OpenMined/PySyft/}{the 117 contributors to PySyft}.

\bibliographystyle{plainnat}
\bibliography{paper}

\end{document}

%% file: sections/introduction.tex
\section{Introduction}

Secure Multiparty Computation (SMPC) is becoming increasingly popular as a way to perform operations in an untrusted environment without disclosing data. In the case of machine learning models, SMPC would protect the model weights while allowing multiple worker nodes to take part in the training phase with their own datasets, a process known as Federated Learning (FL). However, it has been shown that securely trained models are still vulnerable to reverse-engineering attacks that can extract sensitive information about the datasets directly from the model. Another set of methods, labelled as Differentially Private (DP) methods, address this and can efficiently protect the data.

We provide a transparent framework for privacy preserving deep learning to every PyTorch user, enabling the use of FL, MPC, and DP from an intuitive interface. We show the ability of the framework to support various implementations of MPC and DP solutions and report the results obtained when instantiating the SPDZ\cite{10.1007/978-3-642-32009-5_38} and moment accountant\citep{Abadi:2016:DLD:2976749.2978318} methods respectively for MPC and DP in a federated learning context.

Our main contributions are the following:\\
- We first build a standardized protocol to communicate between workers which made federated learning possible.\\
- Then, we develop a chain abstraction model on tensors to efficiently override operations (or encode new ones) such as sending/sharing a tensor between workers.\\
- Last, we provide the elements to implement recently proposed differential privacy and multiparty computation protocols using this new framework.

By doing so, we intend to help popularize privacy preserving techniques in machine learning by making them available via the common tools that researchers and data scientists work with on a daily basis. Our framework is designed in a extensible way such that new FL, MPC, or DP methods can be plugged in by external contributors willing to make their work available to the wider deep learning community.

%% file: sections/framework.tex
\section{A standardized framework to abstract operations on Tensors} \sectionmark{A standardized framework}
\label{sec:stdframework}

%
%
%

\subsection{The chain structure}

Performing transformations or sending tensors to other workers can be represented as a chain of operations, and each operation is embodied by a special class. To achieve this, we created an abstraction called the \textit{SyftTensor}. SyftTensors are meant to represent a state or transformation of the data and can be chained together. The chain structure always has at its head the PyTorch tensor, and the transformations or states embodied by the SyftTensors are accessed downward using the \textit{child} attribute and upward using the \textit{parent} attribute.

\begin{figure}[]
   \begin{minipage}{0.25\textwidth}
     \centering
     \includegraphics[scale = 0.35]{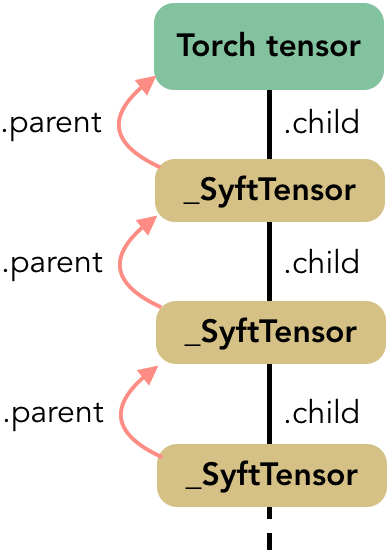}
     \caption{General structure of a tensor chain}
     \label{fig:syft_chain}
   \end{minipage}\hfill
   \begin{minipage}{0.4\textwidth}
     \centering
     \includegraphics[scale = 0.30]{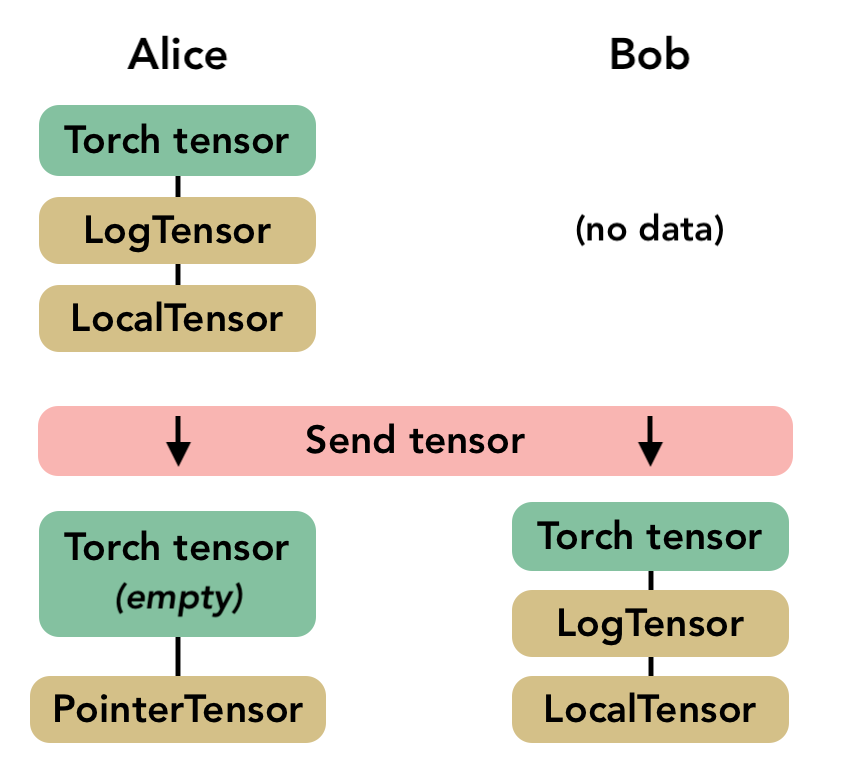}
    \caption{Impact of sending a tensor on the local and remote chains}
    \label{fig:chain_sent}
   \end{minipage}
   \begin{minipage}{0.3\textwidth}
     \centering
     \includegraphics[scale=0.3]{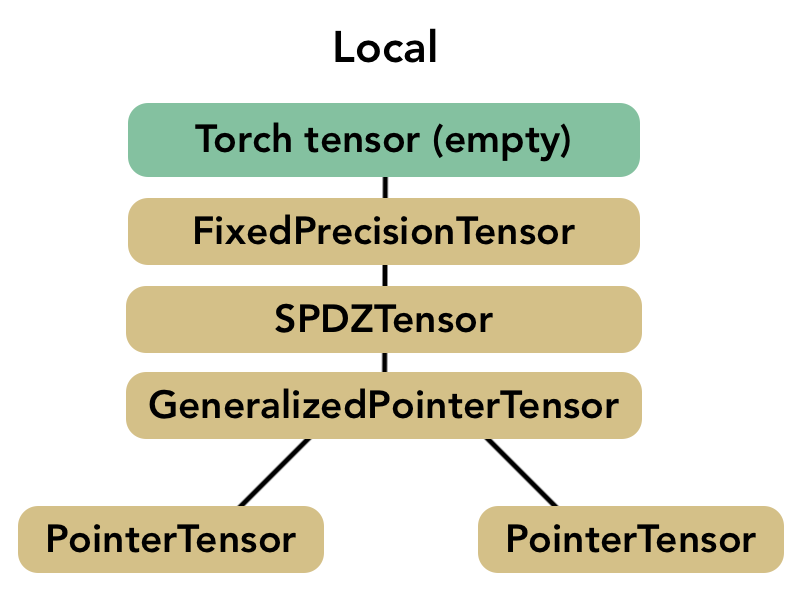}
     \caption{Chain structure of a SPDZ tensor}
     \label{fig:mpc_chain}
   \end{minipage}
\end{figure}

Figure \ref{fig:syft_chain} presents the general structure of a tensor chain, where SyftTensors are replaced with instances of some subclasses which all have a specific role, like the \texttt{LocalTensor} class which will be described next. All operations are first applied to the Torch tensor which makes it possible to have the native Torch interface, and they are then transmitted through the chain by being forwarded to the child attribute.

There are two important subclasses of SyftTensor. First, the LocalTensor which is created automatically when the Torch tensor is instantiated. Its role is to perform on the Torch tensor the native operation corresponding to the overloaded operation. For instance, if the command is \texttt{add}, then the LocalTensor will perform the native torch command \texttt{native\_add} on the head tensor. The chain has two nodes and it loops so that the LocalTensor child refers to the head node tensor which contains the data without needing the re-create a child tensor object, which would reduce performance.

Second, the PointerTensor which is created when a tensor is sent to a remote worker. Sending and getting back a tensor is as simple as calling the methods \texttt{send(worker)} and \texttt{get()} on the tensor. When this happens, the whole chain is sent to the worker and replaced by a two-node chain: the tensor, now empty, and the PointerTensor which specifies who owns the data and the remote storage location. This time, the pointer has no child. Figure \ref{fig:chain_sent} illustrates how the chains are modified when being sent to a remote worker and how LocalTensor and PointerTensor are used in those chains.


\subsection{From virtual to real context execution of federated learning}

In order to simplify debugging complex chains of operations, this framework develops the notion of \textit{Virtual Workers}. Virtual Workers all live on the same machine and do not communicate over the network. They simply replicate the chain of commands and expose the very same interface as the actual workers to communicate with each other.

Network workers in the context of Federated Learning have two implementations in the framework as of now. One builds upon plain network sockets, while the other supports Web Sockets. Web Socket workers allow multiple workers to be instantiated from within a browser, each within its own tab. This gives us another level of granularity when building federated learning applications before actually addressing remote workers not on the same machine. Web Socket workers are also a very good fit for the data science ecosystem revolving around browser-based notebooks.

%% file: sections/mpctensor.tex
\section{Towards a Secure MPC framework} \label{sec:mpctensor}

\subsection{Building an MPCTensor}

The elements introduced in Section \ref{sec:stdframework} form the building bricks necessary to create our \textit{MPCTensor}. Splitting and sending the shares can be done using a list of \textit{PointerTensors}. The MPC toolbox proposed in our framework implements the SPDZ protocol from \cite{damgaard2013practical,10.1007/978-3-642-32009-5_38}.

The MPC toolbox includes basic operations such as addition and multiplication but also preprocessing tools to generate for instance triples used for multiplication, and more specific operations to neural networks including matrix multiplication. Some adjustments are made to the traditional elements of a convolutional network due to the specificities of MPC: as described in \cite{10.1007/978-3-642-32009-5_38}, we use average pooling instead of max pooling and approximate higher-degree sigmoid instead of relu as an activation function. 

Since the SPDZ protocol assumes that the data is given as integers, we added into the chain a node called the \textit{FixedPrecisionTensor} that converts float numbers into fixed precision numbers. This node encodes the value into an integer and stores the position of the radix point. The complete structure of a tensor implementing SPDZ is summarized in figure \ref{fig:mpc_chain}.


Unlike the MPC protocol proposed by \cite{10.1007/978-3-642-32009-5_38}, players are not equal in our framework since one is the owner of the model (called the \textit{local worker}). He acts as a leader by controlling the training procedure on all the other players (the \textit{remote workers}). To mitigate this centralization bias when dealing with data, the local worker can create remote shared tensors on data he doesn't own and can't see.


Indeed, we expect remote workers to hold some data of their own in a general setting, for instance when hospitals are contributing medical images to train a model. Multiple players are then interested in seeing the execution performing correctly, which is particularly crucial during the inference phase where many factors could lead to corrupted predictions \cite{ghodsi2017safetynets}.

So far, the current implementation does not come with a mechanism to ensure that every player behaves honestly. An interesting improvement would be to implement MAC authentication of the secret shared value, as suggested by \cite{10.1007/978-3-642-32009-5_38}.

\subsection{Applying Differential Privacy}

We implemented differential privacy based on the work of \cite{Abadi:2016:DLD:2976749.2978318}, which provides a training method for deep neural networks within a modest ("single-digit") privacy budget. To achieve this, the paper provides a new estimate of the privacy loss used to carefully adjust the noise needed, along with a new algorithm improving the efficiency of the private training.

In particular, we implemented Stochastic Gradient Descent (SGD): instead of iterating in the same way over the dataset and over epochs, the training is made of phases, each of them consisting of sampling $L$ items from the $N$ items of the dataset and using them to upgrade the model. We directly reused the \textit{privacy accountant} provided by \cite{Abadi:2016:DLD:2976749.2978318}, but implemented our own \textit{sanitizer} which clips gradients and adds Gaussian noise.

Our framework also provides some refinements guided by the federated learning context. First, when sampling a lot, we randomly choose a worker and sample among its own data. Second, gradients are sanitized on the remote worker in order to efficiently ensure data-privacy. This way, the local worker will get secured gradients for updating the model which cannot disclose information about the dataset.

The approach described in \cite{journals/corr/PapernotAEGT16} proposes another approach to ensure differential privacy by training the final model (called the \textit{student} model) using the noisy and aggregated votes of pre-trained and unpublished models (the \textit{teachers}). It is currently being implemented and will be integrated as another DP Tensor in our framework.

%% file: sections/results.tex
\section{Results and discussion} \sectionmark{Results and discussion}

Table \ref{tab:fed_boston} reports the execution time required to train a neural network on the canonical Boston Housing dataset, using three declinations of our framework. A performance analysis denotes a reasonably small overhead for using Web Socket workers instead of Virtual Workers, thus validating their purpose of notebook-developer tool. This is due to the low network latency when communicating between different local tabs. We are however 46 times slower than using regular PyTorch. We observe the same overhead in performance in our second experiment that trains a classifier to detect diabetes using the Pima Indian Diabetes dataset, a small dataset containing 768 rows and 8 columns \cite{dat:PIMA}. 

Table \ref{tab:dp_boston} shows how increasing $\epsilon$ improves the model at the expense of data privacy. The DP model achieves a 25-30 MSE compared to 20-24 in the baseline model, but the privacy guarantee remains strong as we achieve (0.5, $10^{-5}$)-differential privacy. These results are consistent with those reported in the literature for computer vision applications \cite{Abadi:2016:DLD:2976749.2978318}.

\begin{table}
    \parbox{.5\linewidth}{
    \centering
    \begin{tabular}{|l|c|}
      \hline
      Training mode & Training time (s) \\
      \hline
      PySyft (Virtual) & 10.1 \\
      PySyft (Socket) & 14.6 \\
      PySyft (Virtual) + DP* & 15.3 \\
      Pure PyTorch & 0.22 \\
      \hline
    \end{tabular}
    \caption{Training time using different training settings on the Boston Housing dataset (10 epochs) \\\textit{*Equivalent time for the same number of batches processed for DP}}
    \label{tab:fed_boston} 
    }
    \hfill
    \parbox{.45\linewidth}{
    \centering
    \begin{tabular}{|l|c|c|}
          \hline
          ($\epsilon$, $\delta$)-privacy & Boston MSE & Pima Acc. \\
          \hline
          (0.5, $10^{-5}$) & 29.4 & 60.6\%\\
          (1, $10^{-5}$) & 29.2 & 64.2\%\\
          (2, $10^{-5}$) & 28.5 & 66.1\%\\
          (4, $10^{-5}$) & 28.6 & 67.1\%\\
          \textit{no privacy} & 23.7 & 70.3\% \\
          \hline
        \end{tabular}
        \caption{Accuracy of differentially private federated learning on the Boston Housing and Pima Diabetes datasets}
        \label{tab:dp_boston} 
    }
\end{table}

For the Boston Housing dataset, the baseline model spends approximately 19.8ms per batch while the differentially private model spends about 30.0ms, which is a very reasonable overhead ($+50\%$) for a feature like privacy. One last observation that we can make is that the convergence is far slower with DP enabled. The MSE keeps a value in the range of 500 over a first phase of 50 samplings. Then the MSE starts decreasing and steadily reaches a 10-50 MSE value. Two reasons can explain this behaviour: first, gradient clipping reduces the efficiency of the updates from the last layers, and second the Gaussian noise interferes with the updates suggested by the gradients which are therefore less efficient. Note that raising the bound for gradient clipping also increases the variance of the Gaussian noise.

%% file: sections/conclusion.tex
\section{Conclusions}
\label{conclusions}
\vspace{-2px}
We have introduced a privacy preserving federated learning framework built over PyTorch. The design relies on chains of tensors that are exchanged between local and remote workers. Our tensor implementations support commands of the PyTorch API and combine MPC and DP functionalities within the same framework.

There are still many issues to address, at the forefront of which is decreasing training time. Efficiency has not been tackled yet, but the current overhead suggests that there is room for improvement in what is a pure Python framework as opposed to high-level Python APIs that piggyback on optimised low-level libraries. Another concern has to do with securing MPC to make sure to detect and defeat malicious attempts to corrupt the data or the model. 

All code samples involved in this paper will be made available in a GitHub repository after satisfying the anonymisation requirements of the submission.

